\documentclass[lettersize,journal]{IEEEtran}
\usepackage{amsmath,amsfonts}
\usepackage{algorithmic}
\usepackage{algorithm}
\usepackage{array}
\usepackage[caption=false,font=normalsize,labelfont=sf,textfont=sf]{subfig}
\usepackage{textcomp}
\usepackage{stfloats}
\usepackage{url}
\usepackage{verbatim}
\usepackage{graphicx}
\usepackage{cite}
\usepackage{multirow}
\usepackage[table]{xcolor}
\usepackage[capitalize]{cleveref}
\usepackage{pifont}
\begin{document}

\title{Two-Stage Prompt Learning: A Unified Framework for Person ReID without Clothing Hypothesis}

\author{Yuxuan~Xiao, Shanshan~Zhang,~\IEEEmembership{Member,~IEEE}
                Shengcai Liao,~\IEEEmembership{Fellow,~IEEE}
                and Jian~Yang,~\IEEEmembership{Member,~IEEE}
        % <-this % stops a space

\thanks{\emph{Corresponding author}: Shanshan Zhang.} % <-this % stops a space
\thanks{Y. Xiao, J. Yang and S. Zhang are with the PCA Lab, School of Computer Science and Engineering, Nanjing University of Science and Technology, Nanjing 210094, China (e-mail: xiaoyuxuan@njust.edu.cn; csjyang@njust.edu.cn; shanshan.zhang@njust.edu.cn).} % <-this % stops a space
\thanks{S. Liao is with the College of Computing and Artificial Intelligence, United Arab Emirates University, Al Ain, UAE (e-mail: scliao@ieee.org).}
\thanks{This work was supported in part by the National Natural Science Foundation of China (Grant No. 62322602), and in part by the Natural Science Foundation of Jiangsu Province, China (Grant No. BK20230033).}}
% \thanks{Manuscript received Jan, 2025.}}

% \markboth{IEEE Transactions on Information Forensics and Security, ~Vol.~X, No.~X, August~2021}%
% {Shell \MakeLowercase{\textit{Xiao et al.}}: A Coherent Framework for Person Re-Identification under Arbitrary Clothes Conditions via Dual Prompts}


\maketitle

\begin{abstract}
Person re-identification (ReID) in real-world scenarios involves three clothes conditions: the same clothes (SC), changed clothes (CC), and the unknown (GEN) scenario where clothes changes are not specified. Existing methods typically train separate SC- or CC-specific models, which limits generalization across all conditions. Meanwhile, pre-trained visual–language models (VLMs) such as CLIP have shown strong generalization ability in various vision tasks, motivating recent ReID methods to incorporate textual prompts. However, to distinguish SC and CC, prior approaches rely on fixed predefined instructions, which lack learnable context embeddings in prompts and cannot effectively handle the GEN scenario. To overcome these limitations, we propose \textbf{Two-Stage Prompt Learning (TSPL)}, a unified framework that integrates \textit{clothes–identity decoupling} and \textit{scenario-aware task prompts} within VLMs. In \textbf{Stage 1}, appearance-decomposed prompt learning disentangles clothes and identity information under frozen encoders, producing stable embeddings where clothes cues can be selectively preserved or suppressed. In \textbf{Stage 2}, task-aware prompt learning trains the image encoder with learnable context embeddings guided by Stage-1 representations, enabling robust adaptation across SC, CC, and GEN conditions. Extensive experiments on five benchmarks show that TSPL achieves state-of-the-art performance under all scenarios, 
% improving mAP by up to 9.2\% on LTCC (SC) and by 6.6\% and 5.1\% on VC-Clothes (CC and GEN), respectively.
improving mAP by up to 9.2\% on LTCC (SC) and by 6.4\% on VC-Clothes (GEN), respectively. The code will be released at \url{https://github.com/YuSuen/TSPL}.
\end{abstract}


\section{Introduction}

\IEEEPARstart{P}{erson} re-identification (ReID) aims to identify the same individual across images captured by different, non-overlapping cameras. Traditional ReID methods~\cite{Jaha2016, zhu2017image,zhou2019omni,Luo2020,yan2021beyond,he2021transreid,li2023clip,jia2025disaggregation} assume that each person always wears the same clothes (SC) across cameras. While this assumption holds in short-term settings, it often fails in long-term or real-world scenarios where people may change clothes. Consequently, recent studies~\cite{Gu2022,Cui2023, han2023clothing,yang2023good,liu2023learning} have focused on the changed clothes (CC) scenario by learning clothes-invariant features. However, suppressing clothes cues for CC often compromises performance under SC, where clothes information is actually beneficial. 
This inherent contradiction limits the adaptability of existing frameworks across diverse scenarios, especially when the clothing condition is unknown. In practical applications such as surveillance-based suspect retrieval, it is often unclear whether a person has changed clothes or not: in many short-term cases, individuals typically remain in the same outfit, while in other situations, they may intentionally change clothes to evade tracking or appear under different clothing conditions due to time gaps (e.g., entering and leaving a building).
Therefore, there is a pressing need for a unified framework that can effectively handle \textit{all} clothes conditions, including SC, CC, and the unknown (GEN) scenario, where it is unclear whether clothes has changed.

\begin{figure}
\begin{center}
  \includegraphics[width=1.0\linewidth]{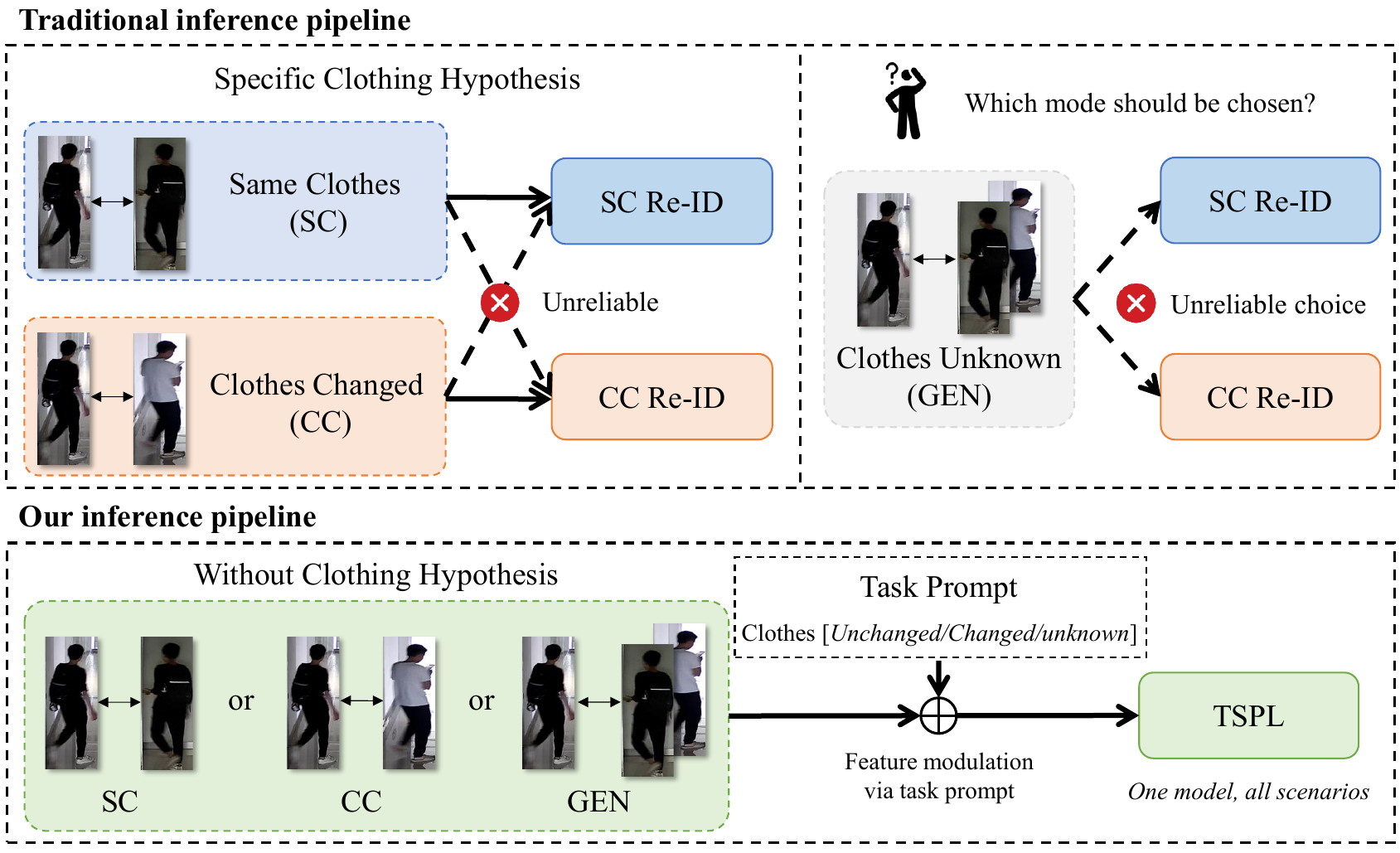}
\end{center}
  \caption{
      Comparison between the traditional inference pipeline based on specific clothing hypothesis and our unified inference framework without clothing hypothesis. In traditional methods, SC and CC are treated as separate modes, requiring prior knowledge of clothes conditions; incorrect mode selection leads to degraded performance, and the GEN scenario introduces ambiguity in choosing between SC and CC models. In contrast, our method (bottom) adopts a unified pipeline, where image pairs under different clothes conditions (SC, CC, or GEN) are directly processed with a task prompt and fed into the proposed TSPL, eliminating the need for explicit mode selection and enabling robust inference across all scenarios.
      % Comparison of existing methods under different clothes conditions. Red boxes indicate misidentifications, and R@1 denotes Rank-1 accuracy. SC-specific methods fail under CC due to over-reliance on clothes cues, while CC-specific methods discard clothes information, limiting SC performance. Our method unifies both strategies, achieving robustness across SC and CC.
  }
\label{fig:intro}
\end{figure}

Meanwhile, pre-trained visual–language models (VLMs) such as CLIP~\cite{radford2021learning} have demonstrated remarkable generalization ability in various vision tasks through prompt learning~\cite{zhou2022learning,zhou2022conditional,vidit2023clip,wang2023all,lee2023multimodal,yu2025clip}. However, existing prompt-based ReID approaches remain limited. CLIP-ReID~\cite{li2023clip} only learns identity prompts while ignoring clothes variations, leading to poor performance under CC. Instruct-ReID~\cite{he2023instructreid} introduces task instructions to guide feature learning under SC and CC. However, it is not a prompt learning method, because its instructions only consist of predefined words (e.g., ``change clothes'' or ``do not change clothes'') without learnable context-aware prompt embeddings, limiting its adaptability and generalization, especially to the GEN scenario. These limitations motivate us to develop a more flexible and generalized prompt learning framework.

We propose \textbf{Two-Stage Prompt Learning (TSPL)} to address these challenges. As summarized in Table~\ref{tab:method_comparison}, TSPL integrates \textit{clothes–identity decoupling} and \textit{scenario-aware task prompts} into a unified two-stage prompt learning framework. In \textbf{Stage 1}, we freeze the image encoder and align the text encoder with its output to learn disentangled clothes and identity prompt embeddings via pseudo-words. This appearance-decomposed learning produces stable embeddings where clothes cues can be selectively preserved or suppressed, which will later guide Stage 2 training.
In \textbf{Stage 2}, the image encoder is unfrozen and trained with scenario-aware task prompts, where a learnable context prompt embedding is introduced for task-specific adaptation. The disentangled embeddings from Stage 1 are selectively activated based on SC/CC instructions to guide image encoder learning, while GEN is handled at inference via masking these instructions.

\begin{table*}[t]
\centering
\caption{Comparison between CLIP-ReID~\cite{li2023clip}, Instruct-ReID~\cite{he2023instructreid}, and our proposed TSPL method. TSPL integrates appearance–identity decoupling and learnable, scenario-aware task prompts, enabling robust performance across all clothing conditions.}
\label{tab:method_comparison}
\begin{tabular}{lccccc}
\hline
Method                              & ID Prompt                     & Clothes Prompt              & Task Prompt                              & Clothes Scenarios        \\ \hline
CLIP-ReID~\cite{li2023clip}         & Learnable (ID only)           & \ding{55} No                & \ding{55} No                             & SC only                  \\
Instruct-ReID~\cite{he2023instructreid}    & \ding{55} No           & \ding{55} No                & Fixed instruction words (non-learnable)  & SC + CC only               \\
\textbf{TSPL (Ours)}                & Learnable (ID + Clothes)      & Decoupled, Learnable        & \ding{51} Learnable task prompts         & SC + CC + GEN             \\ \hline
\end{tabular}
\end{table*}

The two-stage design is essential: Stage 1 provides disentangled representations under fixed encoders, while Stage 2 uses them to guide unified feature learning across scenarios. A single-stage approach cannot achieve such disentangled and scenario-aware modeling. 

Overall, TSPL differs fundamentally from CLIP-ReID and Instruct-ReID: it handles all clothes conditions (SC, CC, GEN), learns both appearance and task prompts rather than using fixed instructions, and explicitly decouples clothes and identity information. As a result, TSPL achieves robust and generalized performance across diverse ReID scenarios rather than being a straightforward combination of existing methods.

\textbf{Key Contributions.} 
This work makes three main contributions: 
(1) We introduce a unified prompt learning framework that explicitly decouples clothes and identity information, enabling a single model to handle all clothes conditions (SC, CC, GEN). 
(2) We design a two-stage learning strategy that first learns stable, disentangled prompt embeddings with frozen VLM encoders, and then uses them to guide scenario-aware prompt learning with a trainable image encoder for robust generalization. 
(3) Extensive experiments on five benchmark datasets demonstrate that our method consistently outperforms prior approaches, achieving state-of-the-art results across all scenarios, with improvements of up to 9.2\% in mAP under SC on LTCC and 6.6\% and 5.1\% under CC and GEN on VC-Clothes, respectively.

\section{Related Work}
\label{sec:relatedwork}

\subsection{Person Re-identification}
Person re-identification (ReID)~\cite{liu2023weakly,liu2024two,liu2025try, wei2025multiple} aims to match images of the same person across different non-overlapping cameras.
Traditional ReID methods~\cite{Jaha2016, zhou2019omni, Luo2020, he2021transreid, li2023clip} are based on the assumption that the clothes of the same person is invariant. However, this assumption does not always hold in practice especially for long-term ReID, limiting the applications of these methods. Therefore, growing interest emerges in clothes changed ReID. 
Some of existing clothes changed ReID methods aim to seeking some easily available clothes-irrelevant clues such as radio signals~\cite{fan2020learning}, silhouettes~\cite{hong2021fine}, contour sketches~\cite{yang2019person}, skeletons~\cite{qian2020long}, 3D shape~\cite{Chen2021, liu2023learning}, human body semantic maps~\cite{guo2023semantic} and gait~\cite{jin2022cloth}. 
Recently, performance has been largely boosted by using dense 2D-3D correspondences~\cite{wang2023exploring}, which are yet hard to obtain in practice.

For the sake of efficiency and convenience, some other methods propose to extract clothes-irrelevant features directly from RGB images without other cues. 
% RCSANet~\cite{huang2021clothing} utilizes appearance features to guide feature clustering of the same identity, allowing differentiation between different identities in high-dimensional space; 
For example, CAL~\cite{Gu2022} proposes a clothes based adversarial loss function to penalize the discriminative power of the clothes classification model, thereby extracting clothes-irrelevant features;
DCR-ReID~\cite{Cui2023} extends CAL by decoupling clothes-relevant and clothes-irrelevant features based on human body region and contour features;
AFD-Net~\cite{Xu2021} uses intra-class reconstruction and inter-class adversarial learning to disentangle identity-related and identity-unrelated features;
CCFA~\cite{han2023clothing} and AIM~\cite{yang2023good}improves the robustness to clothes variations by clothes augmentation in the feature space.

However, existing CC ReID methods do not perform well under the SC scenario, as they get rid of clothes-relevant information, which is the most effective cue for SC ReID. In contrast, our method is more compatible to different clothes conditions, including CC, SC and GEN.

\begin{figure*}[ht]
\begin{center}
  \includegraphics[width=0.94\linewidth]{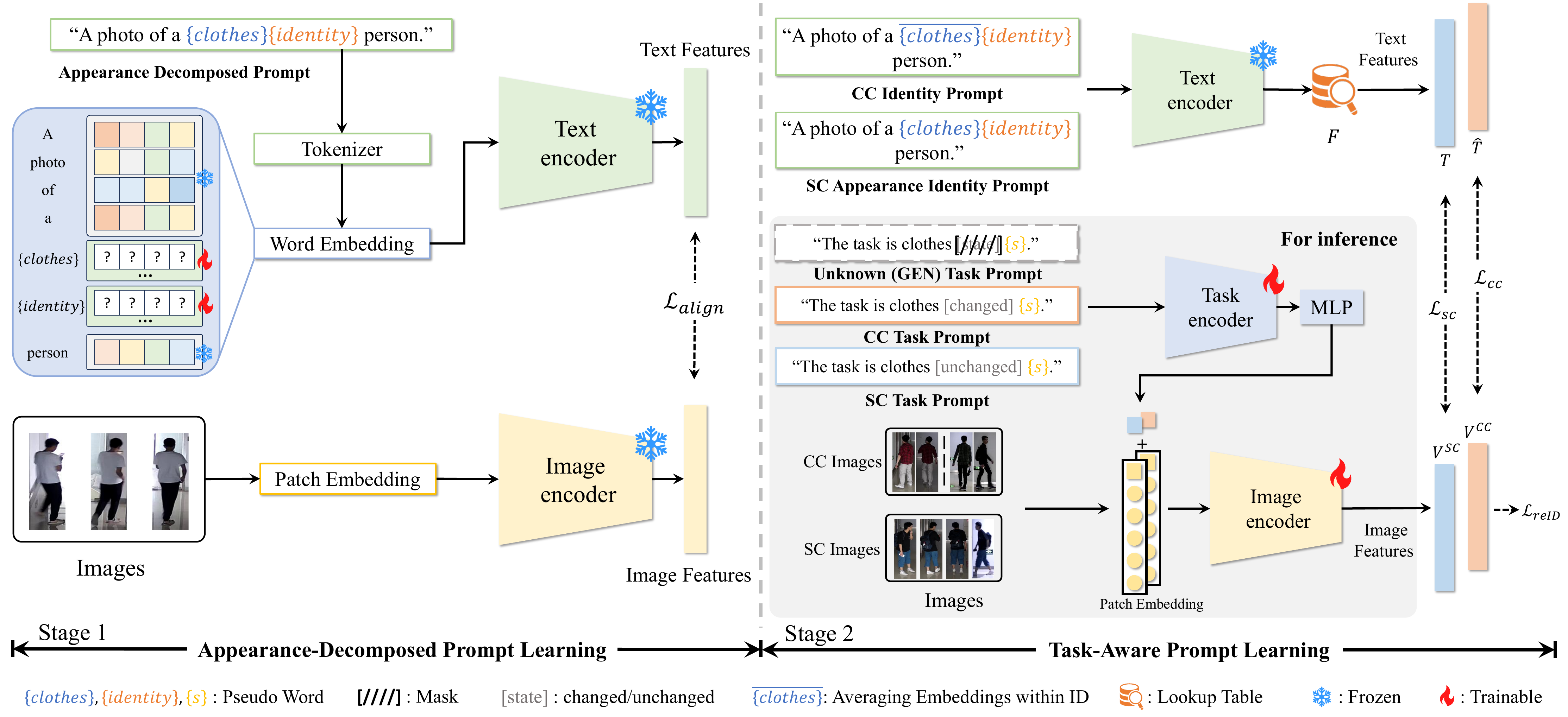}
\end{center}
    \caption{
    Overview of our proposed method, which consists of two stages: appearance-decomposed prompt learning and task-aware prompt learning.
    \textbf{Stage 1:} we design an appearance-decomposed prompt to separately encode clothes-relevant and identity-relevant information into distinct pseudo-word embeddings.
    \textbf{Stage 2:} we choose to either suppress or preserve clothes-relevant information based on the learned appearance-decomposed prompts according to the specific scenario, and then use the text prompts to guide customized image feature extraction.
    Furthermore, we introduce a task-aware prompt learning with a self-adaptive pseudo-word embedding to unify diverse feature extraction strategies within a unified framework. Note that GEN prompt is only used for inference.
    During inference, only the task encoder and image encoder are used to produce scenario-specific features for retrieval.
    }
  % \caption{Overview of our proposed method, which comprises two stages: appearance decomposed prompt learning and task specific prompt learning guided joint training. 
  % At the first stage, we propose an appearance decomposed prompt to incorporate clothes-relevant and identity-relevant information into different pseudo-word embedding entries, respectively.
  % At the second stage, we choose to either suppress or preserve clothes-relevant information based on the learned appearance decomposed prompts according to the specific scenario, and then use the text prompts to guide customized image feature extraction.
  % Furthermore, we introduce a task specific prompt with self-adapted pseudo-word embedding to integrate different feature extraction strategies into a coherent framework, so as to enhance the efficiency. For inference, we only use the task encoder and image encoder to generate customized features for retrieval, according to the specific clothes condition.}
  \label{fig:overview}
\end{figure*}

\subsection{Prompt Learning for ReID}

CLIP-ReID~\cite{li2023clip} first introduces prompt learning to the ReID task. It proposes to learn a common prompt embedding associated with pseudo words for each identity as ID descriptions, which are then used to align image features during contrastive learning. 
The fine-grained attribute descriptions can significantly supplement the valuable semantic information for the person images. Instead of learning ID prompts, MP-ReID~\cite{zhai2024multi} uses ChatGPT, VQA and implicit learning to generate multiple prompts for person images, and then aligns them with image features through cross-modal alignment.
However, CLIP-ReID and MP-ReID focus on the traditional ReID task, i.e. the SC scenario in this work and do not consider the CC scenario. 

Several recent works, including CCAF~\cite{li2024clip}, SCI~\cite{han2024see}, and MIPL~\cite{wei2025multiple}, adopt an explicit image-level disentanglement strategy. These methods decompose images into clothes-related and identity-related components, learn corresponding prompts, and perform feature alignment to extract clothes-irrelevant features. DIFFER~\cite{liang2025differ} models appearance, pose, and biometric information at the textual level with the help of an external VLM, and applies gradient reversal to suppress non-biometric factors for clothes invariant feature learning.
However, these methods focus exclusively on learning clothes-irrelevant features for the CC scenario, without considering feature learning for the SC and GEN scenario.

Recently, Instruct-ReID~\cite{he2023instructreid} integrates different ReID tasks (including traditional ReID, clothes-changing ReID, clothes template based clothes-changing ReID, visible-infrared ReID, text-to-image ReID and language-instructed ReID) into a unified framework by designing different instructions, which are used to modify the image encoder accordingly.
In contrast, our work addresses the unique challenges under varying clothes conditions through prompt learning, instead of relying solely on prompts.
We not only propose that appearance-decomposed prompt learning allow for customized feature extraction under different clothes conditions, but also explore incorporating different customized feature extraction strategies into a unified framework via learnable adaptive task prompt. However, they simply integrate SC, CC, and other ReID task datasets into one framework for training via prompt.
Our method outperforms Instruct-ReID by a large margin (up to 17\% in Tab. \ref{tab:compare-sota}), showing the superiority of handling the special challenge of clothes conditions over a unified framework for wider settings.

\section{Methodology}
\label{sec:methodology}
% In this section, we provide an overview of our method TSPL, followed by detailed descriptions of each stage: appearance decomposed prompt learning (\cref{sec:contrastive_learning}) and task-specific prompt learning guided joint training (\cref{sec:joint_training}). Finally, we explain the inference procedure (\cref{sec:inference}).

\subsection{Overview}
\label{sec:overview}
An overview of our method TSPL is shown in~\cref{fig:overview}.
Our method is built upon the CLIP~\cite{radford2021learning} model, whose image encoder $\mathcal{I}(\cdot)$ and text encoder $\mathcal{T}(\cdot)$ are used for feature extraction.
% The entire method consists of two stages: appearance-decomposed prompt learning and task-aware prompt learning guided joint training. 
The entire method consists of two stages: appearance-decomposed prompt learning and task-aware prompt learning. 

For appearance-decomposed prompt learning, we develop a learnable appearance-decomposed prompt (ADP), in embedding space which includes a set of clothes-relevant learnable vectors and another set of identity-relevant learnable vectors. 
By applying CLIP-style contrastive learning, the appearance-decomposed prompt helps us to obtain decoupled clothes-relevant and identity-relevant information in learnable vectors, being detailed in~\cref{sec:contrastive_learning}.

For task-aware prompt learning, we consider preferences across different scenarios. Based on the appearance-decomposed prompts from the first stage, we choose different combinations of clothes-relevant and identity-relevant prompt information for each identity depending on the specific scenario. The corresponding prompt text features are then used as guidance for customized image feature extraction via alignment between text and image features.
To integrate different feature extraction strategies into one unified framework, we introduce a task-aware prompt (TAP) to modulate image features. This prompt allows scenario-specific image features to be controlled in a unified framework by adjusting task prompts. See more details in~\cref{sec:joint_training}.

% Considering the preferences of different scenarios, we choose different combinations of clothes-relevant and identity-relevant information for each identity depending on the specific scenario. The corresponding text features are used as guidance for customized image feature extraction via alignment between text and image features.
% Furthermore, we introduce a task specific prompt (TSP) to integrate different feature extraction strategies into one coherent framework.
% We first design text descriptions to different clothes conditions, and then inject the text embedding from the task encoder into the [CLS] token~\cite{ma2022multimodal, lee2023multimodal}. Since the output [CLS] token serves as a global feature representation~\cite{li2023clip, he2021transreid}, we can adjust image feature extraction by changing the task specific prompt, see more details in~\cref{sec:joint_training}.
% For inference, we only use the task encoder and image encoder to generate customized features for retrieval, according to the specific clothes condition, see more details in~\cref{sec:inference}.

Overall, the appearance-decomposed prompt learning allows us to disentangle identity-relevant and clothes-relevant information so that we can dynamically adjust the features for different scenarios, for example, under CC, the clothes-relevant information should be suppressed. The task-aware prompt learning allows us to handle multiple clothes conditions (CC, SC and GEN) with a shared feature extractor instead of having separate branches, which makes full use of data under different scenarios to optimize the feature extractor jointly. The above two kinds of prompts complement with each other so that we can handle different scenarios in a unified framework.
For inference, we only use the task encoder and image encoder to generate customized features for retrieval, according to the specific clothes condition, see more details in~\cref{sec:inference}.

\subsection{Appearance-Decomposed Prompt Learning}
\label{sec:contrastive_learning}
As discussed before, SC ReID benefits from clothes-relevant features while CC ReID should focus on other identity-relevant features. 
Therefore, a straightforward way is to decouple the entire appearance features into clothes-relevant and identity-relevant ones, so that we can adjust features under different clothes conditions.
However, existing decoupling methods operate directly in the image~\cite{guo2023semantic} or feature space~\cite{Eom2022, han2023clothing, yang2023good}, which results in breaking down the original semantic structure of the image.
Inspired by CLIP-ReID~\cite{li2023clip}, which learns an ambiguous identity text description for each identity from multiple instances through text-image feature alignment, we propose incorporating clothes variations into identity text description learning to decompose the ambiguous information into specific clothes and identity, thereby avoiding the information corruption caused by image decomposition.
This disentangled text description on clothes-relevant and identity-relevant information is then used for aiding customized feature learning for different clothes scenarios.

To this end, we introduce appearance-decomposed prompt learning. The appearance-decomposed prompt decomposes the description of each person instance into clothes and identity, and then add prefixes and suffixes to form a complete prompt. 
Specifically, our appearance-decomposed prompt contain 4 parts: prefix ``a photo of a'', suffix ``person'', and the pseudo-words \{clothes\} and \{identity\}. 
The design of the prefix and suffix follows the common prompt templates used in CLIP and CLIP-ReID, aiming to maintain semantic stability and reduce variance introduced by free-form textual descriptions.
Note that the pseudo-words here, corresponding to clothes and identity, respectively, do not have any meaning but can be processed together with other words.
To find pseudo-words for specific ID and clothes, we replace the word embeddings of the pseudo-words \{clothes\} and \{identity\} in the text embedding space with $N_{ID}$ set of $u$ learnable clothes context vectors $\{c_1,c_2,\cdots,c_u\}$ and  $N_{CLT}$ set of $v$ learnable identity context vectors $\{h_1,h_2,\cdots,h_v\}$, respectively. Here, $N_{ID}$ and $N_{CLT}$ indicate the number of identities and clothes, respectively; $u$ and $v$ indicate the number of learnable context vectors for clothes and identity respectively. 
This decomposed learnable prompt structure allows us to obtain separable clothes-relevant and identity-relevant information in the text embedding space. 

Formally, for a mini-batch $\left\{{(x_{i},y_{i},z_{i})|^{B}_{i=1}}\right\}$, where $x_i$, $y_{i}$ and $z_{i}$ represent image, identity label and clothes label, respectively, 
we denote the prompt text embedding for an arbitrary image $x_{i}$ as $P_{y_{i}z_{i}}$= \{$\langle \text{a} \rangle$ $\langle \text{photo} \rangle$ $\langle \text{of} \rangle$ $\langle \text{a} \rangle$ $[c_{1},c_{2},\cdots,c_{u}]^{z_{i}}$ $[h_{1},h_{2},\cdots,h_{v}]^{y_{i}}$ $\langle \text{person} \rangle$\}, where $\langle \cdot \rangle$ indicates word embedding operation.
Then, we obtain the text features $\mathcal{T}(P_{y_{i},z_{i}})$ and image features $\mathcal{I}(x_{i})$ by feeding the prompt text embedding $P_{y_{i},z_{i}}$ and the corresponding image $x_{i}$ to the text encoder $\mathcal{T}(\cdot)$ and the image encoder $\mathcal{I}(\cdot)$ of the frozen CLIP model.
We constrain the alignment of text features and image features through CLIP-style contrastive loss, $\mathcal{L}_{img2text}$ and $\mathcal{L}_{text2img}$, to optimize the learnable clothes context vector and the learnable identity context vector, thereby learning the clothes information and identity information from the image~\cite{li2023clip}. The loss function for feature alignment is written as follows:
\begin{equation}
    \mathcal{L}_{align} = \mathcal{L}_{img2text} + \mathcal{L}_{text2img}.
\label{eq:prompts}
\end{equation}
For detailed definition, please refer to~\cite{radford2021learning,li2023clip}.
Note that the same ID shares the same identity context vectors, while the clothes context vectors of different instances from the same identity differ according to the clothes labels. Only when two instances are from the same identity and with the same clothes, they construct a positive pair during alignment. This enables clothes context vectors to learn information that is clothes-related and identity context vectors to learn information that is identity-related, thus achieving the decoupling goal. For convenience and efficiency, we construct a look-up table
$F=\{T_{y_{i},z_{i}}| i\in [1,N]\}$
to store all the appearance prompt text features $T$ generated by the text encoder. In this way, we can find the corresponding appearance prompt text features $T$ given the identity label $y_{i}$ and clothes label $z_{i}$.

\subsection{Task-Aware Prompt Learning}
\label{sec:joint_training}
CLIP-ReID~\cite{li2023clip} demonstrates that the image encoder acquires the ability to extract prompt-aligned image features through the optimization of image-text alignment.
Therefore, we can customize feature extraction by adjusting the pseudo-word corresponding embeddings. The question that follows is how to incorporate different customized feature extractions into a unified framework to avoid cumbersome multi-branch output design.
Previous method, Instruct-ReID~\cite{he2023instructreid}, introduce task instructions to guide feature learning under SC and CC. However, it is not a prompt learning method because it consists only of predefined instructions and lack a learnable context-aware cue vector, which limits its adaptability and generalization ability, especially in the GEN scenario.
To this end, we introduce task-aware prompt learning, which incorporates customized feature learning and task prompt guided training.

% \textbf{Customized Feature Learning.} \quad 
\subsubsection{Customized Feature Learning}
After appearance-decomposed prompt learning, we obtain the appearance prompt text for each ID with specific clothes.
For the SC scenario, both clothes and identity relevant information are equally important for retrieval, so we use the original appearance text feature $T_{y_{i},z_{i}}$ from the look-up table based on the clothes label and identity label.
For the CC scenario, clothes information should be suppressed as the same identity may change clothes. So we first collect all the appearance prompt text features of the current ID under different clothes according to the identity label, and then use the average feature vector $\hat{T}_{y_{i},z_{i}}$ over different clothes to represent this ID.
% Calculating the mean of features is to capture the commonness across different pieces of clothes. Such commonness would get rid of the individuality of each piece of clothes, e.g. color, while preserving the shape information, which well reflects the body shape.
The averaging operation is adopted as a soft mechanism to reduce clothes-specific bias while retaining identity-consistent information. Compared with directly removing clothes-related components, averaging allows shared structural cues across different clothes to be preserved, leading to more stable representations under CC. Secondly, since variations across prompts of the same identity are mainly confined to the clothes-related pseudo-word components, while the identity-related semantics remain consistent, averaging effectively smooths clothes-specific variations without disrupting identity-relevant information.
In this way, we retain the common information across different clothes for each ID by averaging, while discarding the unique information of a single piece of clothes.

During customized feature learning, to optimize the feature learning of SC and CC simultaneously, we construct a mini-batch in the way that it consists of half SC and half CC scenario data $\mathcal{B}=\left\{\mathcal{B}_{sc}; \mathcal{B}_{cc}\right\}$ with the same amount, i.e. $\mathcal{B}_{sc}=\left\{{(x_{i},y_{i},z_{i})|^{B/2}_{i=1}}\right\}$ and $\mathcal{B}_{cc}=\left\{{(x_{i},y_{i},z_{i})|^{B}_{i=B/2+1}}\right\}$. See~\cref{sec:Implementation} for detailed sampling rules.
After that, we align the SC image feature $V_{i}^{sc}=\mathcal{I}(x_{i}), i\in[1,B/2]$ and the CC image feature $V_{i}^{cc}=\mathcal{I}(x_{i}), i\in[B/2 +1,B]$ with the corresponding appearance prompt text feature $T_{i}=T_{y_{i},z_{i}}$ and $\hat{T}_{i}=\hat{T}_{y_{i},z_{i}}$, respectively.
Therefore, customized feature learning loss can be defined as follows:
\begin{equation}
  \mathcal{L}_{cus} = \sum_{i=1}^{B/2}\mathcal{L}_{sc}(i) + \sum_{i=B/2+1}^{B}\mathcal{L}_{cc}(i),
  \label{eq:cus_loss}
\end{equation}
where
\begin{equation}
  \mathcal{L}_{sc}(i) = -\sum_{k=1}^{N_{sc}}q_{k} \log{\frac{\exp(sim(V_{i}^{sc}, T_{k}))}{\sum_{a}^{N_{sc}}1_{a\neq k}\cdot\exp(sim(V_{i}^{sc}, T_{a}))}},
\end{equation}
\begin{equation}
  \mathcal{L}_{cc}(i) = -\sum_{k=1}^{N_{cc}}q_{k} \log{\frac{\exp(sim(V_{i}^{cc}, \hat{T}_{k}))} {\sum_{a}^{N_{cc}}1_{a\neq k}\cdot\exp(sim(V_{i}^{cc}, 
  \hat{T}_{a}))}},
\end{equation}
where $N_{sc}$ and $N_{cc}$ are the numbers of identities for SC and CC respectively; $sim$ represents inner product similarity calculation; $q_k=(1-\epsilon)\delta_{k,y}+\epsilon/N_{sc(cc)}$ represents the soft label in the target distribution.

% \textbf{Task Prompt Guided Training.} \quad 
\subsubsection{Task Prompt Guided Training}
We define the following task-aware prompt: ``the task is clothes [state] \{$s$\}'', where [state] is set to ``changed'' for the CC scenario and ``unchanged'' for the SC scenario; \{$s$\} is the pseudo-word to be determined. 
Similar to the appearance-decomposed prompt, we represent the text embedding of pseudo-words by a set of $w$ learnable context \{$s_1, s_2,\cdots, s_w$\} in the embedding space.
Since the GEN scenario can be regarded as a comprehensive scenario including both SC and CC, we simply join the SC and CC training data with corresponding task-aware prompts.

During training, after encoding the learnable task-aware prompt using a task encoder $\mathcal{E}(\cdot)$ with three transformation layers, we extract the corresponding prompt feature. This feature is projected into the same space as the image encoder’s [CLS] token via an MLP, added element-wise to the original [CLS] token, and appended to the image patch sequence as input to the image encoder~\cite{ma2022multimodal, lee2023multimodal}.
Consequently, by manipulating the task-aware prompt, we obtain task-adaptive image representations.

Following previous works~\cite{zhou2019omni, Luo2020, he2021transreid, li2023clip}, we incorporate cross-entropy loss $\mathcal{L}_{ce}$ and triplet loss $\mathcal{L}_{tri}$ to optimize the image encoder. The ReID loss is written as:
\begin{equation}
    \mathcal{L}_{ReID} = \lambda\mathcal{L}_{ce} + \mathcal{L}_{tri},
\label{eq:reid_loss}
\end{equation}
where $\lambda$ is a weight balancing two loss values.
During the training data sampling, it may happen that the same ID in the mini-batch belongs to different scenarios. Therefore, for triplet samples, those with different scenarios from the anchor must be excluded from the candidate positive samples.

The overall learning objective of our model is a summation of ReID loss~\cref{eq:reid_loss} and customized feature learning loss~\cref{eq:cus_loss} defined as:
\begin{equation}
  \mathcal{L}_{total} = \mathcal{L}_{reID} + \mathcal{L}_{cus}.
\end{equation}
Note that the text encoder $\mathcal{T(\cdot)}$ is frozen during training, while the image encoder $\mathcal{I(\cdot)}$, the task encoder $\mathcal{E}(\cdot)$ and the learnable context vectors $s$ of the task-aware prompt are optimized.

\subsection{Task Prompt Dynamic Inference}
\label{sec:inference}
During inference, we only utilize the task encoder and the image encoder and use the output [CLS] token as global feature. 
Note that the clothes and ID prompt embeddings learned from the first-stage ADP prompt learning are only used during training of the second-stage TAP prompt learning, where they are used to guide the image encoder learning for the alignment loss, but after that they are no longer needed during inference. 
Benefiting from the integration of SC and CC scenarios in a unified framework with task-aware prompts, the model inherently includes the common knowledge of SC and CC, which is similar to how a multi-task model naturally captures the shared knowledge of multiple tasks~\cite{zhang2021survey}. Therefore, for the inference of the GEN scenario, we simply mask out [state] in the task-aware prompt, making the task prompts irrelevant to neither SC nor CC.
In detail, we adopt a 0–1 masking strategy from NLP~\cite{Yoav}, where the [state] token in the task-aware prompt ``the task is clothes [state] \{s\}'' is masked to form the GEN prompt.
During inference, a binary mask is directly applied to the token embeddings. Specifically, tokens with mask value 1 use their normal token embeddings, while tokens with mask value 0 are replaced with a fixed zero vector. For SC and CC scenarios, the mask is set to all ones, so all tokens, including the pseudo-token [state], are preserved. In contrast, for the GEN scenario, the mask value of the [state] token is set to 0, and its corresponding token embedding is replaced with a zero vector, while all other tokens remain unchanged.
This simple operation effectively removes the task-specific cue from the prompt without altering the overall structure. It is conceptually similar to Word Dropout~\cite{iyyer2015deep}, where selectively masking tokens reduces reliance on potentially noisy or unavailable attributes and improves generalization to unspecified scenarios.
In this way, the task-aware prompt allows users to choose appropriate specific task prompts according to different scenarios to generate scenario customized features for retrieval, thus making the retrieval results more accurate.
Overall, our method is able to handle various scenarios via a single model with the dynamic task-aware prompt.

\section{Experiments}
\label{sec:experiments}

% In this section, we first introduce the datasets, evaluation protocols, and implementation details we use for experiments. After that, we compare with the state-of-the-art methods. Finally, we present ablation studies. 

\subsection{Datasets and Evaluation Protocols}
Widely used clothes changed datasets PRCC~\cite{yang2019person}, LTCC~\cite{qian2020long}, and VC-Clothes~\cite{wan2020person} are utilized for our experiments as they contain SC, CC and GEN scenarios. 
Following previous works\cite{Gu2022,huang2024meta}, we also report the results of our method  under GEN on two large-scale datasets LaST~\cite{shu2021large} and DeepChange~\cite{xu2023deepchange}.
PRCC consists of 33,698 images captured by 3 cameras, including 221 identities, each with two different clothes.
LTCC includes a total of 17,119 images from 152 individuals. Each person was captured by at least two cameras, with the number of clothes ranging from 2 to 14. 
VC-Clothes~\cite{wan2020person} is a virtual dataset synthesized by GTA5. It contains 512 identities of 19,060 images in 4 different scenes with significant clothes changes.
LaST~\cite{shu2021large} consists of 10,862 identities and 228,156 images in total. It is highly diverse, capturing from person ages, scenes, weather, daytime and night.
DeepChange~\cite{xu2023deepchange} contains 178,407 images of 1,121 identities, recorded by a real-world surveillance system using 17 outdoor cameras of varying resolutions.

Following the established evaluation protocol from previous works~\cite{qian2020long, huang2021clothing, Gu2022, Cui2023, yang2023good}, three test modes are defined as follows: 1) CC mode: the same identity wears different clothes in the query and gallery images; 2) SC mode: the same identity always wears the same clothes in the query and gallery images; 3) GEN (Standard) mode: for each identity, some instances in the gallery wear the same clothes with the query, while others may not.
% We employ Cumulative Matching Characteristic Rank-1 (R@1) accuracy and Mean Average Precision (mAP) for evaluation. 
Cumulative Matching Characteristic Rank-1 (R@1) accuracy and Mean Average Precision (mAP) are used for evaluation. 

\begin{table*}
\centering
\caption{Comparison on PRCC and LTCC. ``*'' indicates the results are reproduced using the released source codes. DIFFER$\dagger$ reports the performance of the same architecture as ours.} 
\begin{tabular}{l|c|cc|cc|cc|cc|cc|cc} 
\hline
\multirow{3}{*}{Methods} &\multirow{3}{*}{Size} & \multicolumn{6}{c|}{PRCC}                         & \multicolumn{6}{c}{LTCC}                           \\ 
\cline{3-14}
                 &&\multicolumn{2}{c|}{CC} & \multicolumn{2}{c|}{SC} & \multicolumn{2}{c|}{GEN} & \multicolumn{2}{c|}{CC} & \multicolumn{2}{c}{SC} & \multicolumn{2}{c}{GEN} \\ 
\cline{3-14}
                                    &               &R@1   &mAP   &R@1   &mAP   &R@1   &mAP   &R@1   &mAP   &R@1   &mAP   &R@1   &mAP      \\ 
\hline
% GI-ReID~\cite{jin2022cloth}         &256$\times$128 &37.6  &-     &-     &-     &-     &-     &28.9  &14.2  & -    & -    &73.6  &36.1     \\
% 3DSL~\cite{Chen2021}                &256$\times$128 &51.3  &-     &-     &-     &-     &-     &31.2  &14.8  & -    & -    &-     &-        \\
FSAM~\cite{hong2021fine}            &256$\times$128 &54.5  &-     &98.8  &-     &-     &-     &38.5  &16.2  & -    & -    &73.2  &35.4     \\
AIM~\cite{yang2023good}             &256$\times$128 &54.7  &55.0  &100   &99.9  &- &-  &38.3  &17.0  & -    & -    &76.1  &39.1     \\
Instruct-ReID(STL)~\cite{he2023instructreid}  &256$\times$128 &48.1  &46.0  &-     &-     &-     &-     &-     &-     &-     &-     &66.7  &46.7     \\
RCSANet~\cite{huang2021clothing}    &336$\times$336 &50.2  &48.6  &100   &97.2  &-     &-     &-     &-     & -    & -    &-     &-        \\
3DInvarReID~\cite{liu2023learning}  &384$\times$192 &57.2  &56.5  &-     &-     &-     &-     &40.9  &18.9  & -    & -    &73.2  &35.4     \\
AIM~\cite{yang2023good}             &384$\times$192 &57.9  &58.3  &100   &\textbf{99.9}&- &-  &40.6  &19.1  & -    & -    &76.3  &41.1     \\
CCFA~\cite{han2023clothing}         &384$\times$128 &61.2  &58.4  &99.6  &98.7  &-     &-     &45.3  &22.1  & -    & -    &75.8  &42.5     \\
SCNet~\cite{guo2023semantic}        &384$\times$192 &61.3  &59.9  &100   &97.8  &-     &-     &47.5  &25.5  & -    & -    &76.3  &43.6     \\
AFL~\cite{liu2023clothes}            &384$\times$192&57.4  &56.5  &100   &99.7 &-     &-     &42.1  &18.4  &-     &-     &74.4  &39.1    \\
CLIP3DReID~\cite{liu2024distilling} &384$\times$192 &60.6  &59.3  &-     &-     &-     &-     &42.1  &21.7  &-     &-     &-     &-\\
MCSC~\cite{huang2024meta}           &384$\times$192 &57.8  &57.3  &99.8  &99.8  &-     &-     &42.2  &19.4  &-     &-     &73.9  &40.2     \\
SCI~\cite{han2024see}               &384$\times$192 &62.8  &60.0  &100  &99.4  &-     &-     &45.9  &21.6  &-     &-     &79.9  &44.4     \\
DIFFER$\dagger$~\cite{liang2025differ}&224$\times$224 &-     &-   &-     &-   &-     &-     &36.5  &15.6  &-     &-     &72.6  &35.2     \\
% MADE~\cite{-}                       &224$\times$224 &64.3  &59.1  &100   &98.6  &-     &-     &47.4  &24.4  &-     &-     &84.2  &58.2     \\
% FIR$e^2$~\cite{wang2024exploring}   &384$\times$192 &65.0  &63.1  &100   &99.5  &-     &-     &44.6  &19.1  &-     &-     &75.9  &39.9     \\
\hline
% CAL~\cite{Gu2022}                   &384$\times$192 &55.2  &55.8  &100   &99.8  &-     &-     &40.1  &18.0  &-     &-     &74.2  &40.8     \\
CAL*~\cite{Gu2022}                  &384$\times$192 &54.3  &54.8  &100   &99.8  &99.5  &79.9  &39.8  &17.9  &81.5  &65.7  &73.8  &40.3     \\
% DCR-ReID~\cite{Cui2023}             &384$\times$192 &57.2  &57.4  &100   &99.7  &-     &-     &41.1  &20.4  &-     &-     &76.1  &42.3     \\
DCR-ReID*~\cite{Cui2023}            &384$\times$192 &57.3  &57.9  &100   &99.8  &99.6  &80.6  &42.1  &20.3  &81.5  &66.2  &74.0  &42.1     \\
% DLAW~\cite{liu2023dual}             &384$\times$128 &56.2  &57.1  &98.4  &98.7  &-     &-     &58.0  &35.6  &-     &-     &76.7  &49.1     \\
CLIP-ReID*~\cite{li2023clip}        &384$\times$128 &61.5  &57.5  &100   &99.4  &99.5  &82.2  &36.2  &19.6  &84.7  &66.2  &76.7  &42.2     \\
% AIM~\cite{yang2023good}             &384$\times$192 &57.9  &58.3  &100   &99.9  &-     &-     &40.6  &19.1  &83.5  &66.6  &75.9  &41.7     \\
FIRe\textsuperscript{2}*~\cite{wang2024exploring}&384$\times$192 &65.0  &\textbf{63.1}  &100   &99.5  &99.6   &79.8   &44.6  &19.1  &79.9    &59.4    &75.9  &39.9     \\
\hline
TSPL(Ours)                     &256$\times$128 &67.0  &60.0  &100   &99.5  &\textbf{99.8}  &82.3  &52.0  &26.3  &88.2  &74.2  &81.1  &48.0     \\
TSPL(Ours)                     &384$\times$128 &\textbf{68.0}  &60.0  &\textbf{100}   &99.5  &99.6  &\textbf{82.7}  &\textbf{53.6}  &\textbf{27.2}  &\textbf{89.2}  &\textbf{75.4}  &\textbf{81.3}  &\textbf{49.3}     \\
\hline
\end{tabular}
\label{tab:compare-sota}
\end{table*}

\begin{table}
\centering
\caption{Comparison on VC-Clothes} 
\begin{tabular}{l|cc|cc|cc} 
\hline
\multirow{3}{*}{Methods}           & \multicolumn{6}{c}{VC-Clothes}  \\
\cline{2-7}
                                   &\multicolumn{2}{c|}{CC} & \multicolumn{2}{c|}{SC} & \multicolumn{2}{c}{GEN} \\
\cline{2-7}
                                   &R@1   &mAP   &R@1   &mAP   &R@1   &mAP \\ 
\hline
MDLA~\cite{qian2017multi}          &59.2  &60.8  &94.3  &93.9  &88.9  &76.8    \\
PCB~\cite{sun2018beyond}           &62.0  &62.2  &94.7  &94.3  &87.7  &74.6    \\
Part-Aligned~\cite{suh2018part}    &69.4  &67.3  &93.9  &93.4  &90.5  &79.7    \\
FSAM~\cite{hong2021fine}           &67.3  &81.2  &94.7  &94.8  &-     &-    \\
CAL~\cite{Gu2022}                  &81.4  &81.7  &95.1  &95.3  &92.9  &87.2  \\
AFL~\cite{liu2023clothes}          &82.5  &83.0  &-     &-     &93.9  &88.3 \\
MCSC~\cite{huang2024meta}              &83.3  &83.2  &-     &-     &93.2  &87.9 \\
SCI~\cite{han2024see}          &\textbf{91.7}  &87.5  &-     &-     &95.4  &87.5 \\
\hline
TSPL(Ours)                    &89.8  &\textbf{89.8}  &\textbf{95.7}   &\textbf{96.1}  &\textbf{96.1}  &\textbf{93.0} \\
\hline
\end{tabular}
\label{tab:compare-vc}
% \vspace{-10pt}
\end{table}

\begin{table}[t]
\centering
\caption{Comparison on LaST and DeepChange.} 
\begin{tabular}{l|cc|cc} 
\hline
\multirow{3}{*}{Methods}           & \multicolumn{2}{c|}{LaST}  & \multicolumn{2}{c}{DeepChange}\\
\cline{2-5}
                                   & \multicolumn{2}{c|}{GEN}  & \multicolumn{2}{c}{GEN}\\
\cline{2-5}
                                   &Rank-1   &mAP     &Rank-1   &mAP  \\ 
\hline
BOT~\cite{Luo2020}                 &68.3     &25.3    &47.5     &13.0 \\
mAPLoss~\cite{shu2021large}        &69.9     &27.6    &-        &-    \\
CAL~\cite{Gu2022}                  &73.7     &28.2    &54.0     &19.0    \\
MCSC~\cite{huang2024meta}          &75.4    &31.0     &\textbf{56.9}     &21.5     \\
\hline
TSPL (Ours)                        &\textbf{78.1}  &\textbf{37.2}  &54.5  &\textbf{22.7}  \\
\hline
\end{tabular}
\label{tab:compare-largest}
% \vspace{-15pt}
\end{table}

\subsection{Implementation Details}
\label{sec:Implementation}
We adopt the image encoder (ViT-B/16) and the text encoder from CLIP~\cite{radford2021learning} for our image and text feature extraction.  
Our experiments are conducted on two scales: 256$\times$128 and 382$\times$128.
The hyper-parameters of appearance-decomposed prompt learning are chosen following CLIP-ReID~\cite{li2023clip}.
% For the training process, we apply different sampling rule for each scenario to distinguish SC and CC. 
To distinguish different scenarios during training, we apply different sampling rule. 
For SC, we employ classic \textit{P-K} sampling strategy, so that each mini-batch comprises 64 images from 8 identities, and all instances for the same identity wear the same clothes.
In contrast, for CC, we introduce clothes changes to \textit{P-K} sampling strategy~\cite{Luo2020}. In this case, each mini-batch consists of 64 images from \textit{P}=8 identities, and we make sure each identity has two clothes, each with 4 samples.
We train the model for 45 epochs. We perform a warm-up strategy~\cite{Luo2020} for the model during the initial 10 epochs, gradually increasing the learning rate from $5\times10^{-7}$ to $5\times10^{-6}$. Then, it is decreased by a factor of 0.1 at 15th and 30th epoch respectively.
Adam~\cite{Kingma2015} is adopted as our optimizer. The margin of triplet loss is set as $0.3$. 
The $\lambda$ in~\cref{eq:reid_loss} is set to $0.25$.
Following CLIP-ReID~\cite{li2023clip} and TransReID~\cite{he2021transreid}, the coefficient of Side Information Embedding (SIE) is set to $3.0$. The number of clothes and identity learnable context vectors in appearance-decomposed prompt $u$ and $v$ is set to $4$. The number of the learnable context vectors in task-aware prompt $w$ is set to $4$.

\subsection{Comparison with State-of-the-art Methods}
We compare our method with state-of-the-art methods on PRCC, LTCC, VC-Clothes, LaST and Deepchange. 
As previous methods only present CC/SC numbers on PRCC and CC/GEN numbers on LTCC in their papers, we re-implement some recent methods with released codes (including CLIP-ReID\footnote{\url{https://github.com/Syliz517/CLIP-ReID}}~\cite{li2023clip}, CAL\footnote{\url{https://github.com/guxinqian/Simple-CCReID}}~\cite{Gu2022}, FIRe\textsuperscript{2}\footnote{\url{https://github.com/QizaoWang/FIRe-CCReID}}~\cite{wang2024exploring} and DCR-ReID\footnote{\url{https://github.com/PKU-ICST-MIPL/DCR-ReID_TCSVT2023}}~\cite{Cui2023}), so as to provide more comprehensive comparisons across diverse clothes conditions.
Since there is no GEN evaluation benchmark on PRCC yet, we create a new one as follows: we use the data under camera A on the test set as the query, and take the clothes-consistent data from camera B and the clothes-changed data from camera C as the gallery.

\begin{figure*}
\begin{center}
  \includegraphics[width=1.0\linewidth]{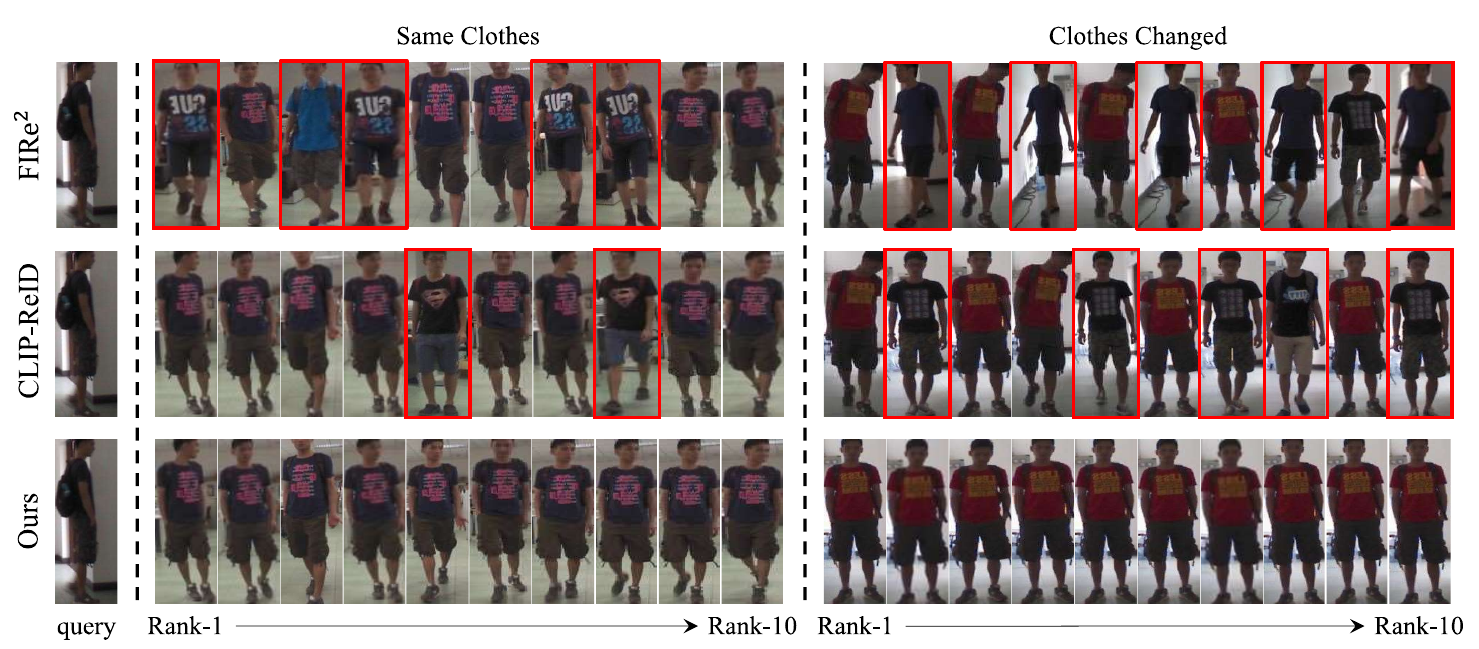}
\end{center}
  \caption{Comparison of qualitative results on PRCC. Red boxes indicate mis-identifications. Compared to other methods, our TSPL performs better under both SC and CC scenarios by a unified model.}
  \label{fig:visual}
\end{figure*}

\textbf{Quantitative Comparison.}\quad
We show comparisons to state-of-the-art methods in~\cref{tab:compare-sota},~\cref{tab:compare-vc} and ~\cref{tab:compare-largest}.
On PRCC, our method significantly outperforms the previous top CC ReID method FIRe\textsuperscript{2} by 3.0\% w.r.t. Rank-1 under CC, and by $\sim$3\% w.r.t. mAP under GEN. Since the SC scenario on PRCC is rather simple, the performance saturates. Our method achieves 100\% w.r.t. Rank-1, comparable to previous methods.
On the more challenging dataset LTCC, our method also significantly outperforms previous methods. 
In particular, our method outperforms the previous top CC ReID method SC-Net by 6.1\% w.r.t. Rank-1 under CC, and by 5.7\% w.r.t. mAP under GEN.
For SC, our method outperforms the previous top SC ReID method CLIP-ReID by 9.2\% w.r.t. mAP as it considers different feature extractions in a unified framework while CLIP-ReID does not.
On the virtual dataset VC-Clothes and two large scale datasets (LaST and Deepchange), we compare the results of TSPL at 384$\times$128 with existing CC ReID methods and achieve state-of-the-art performance.
% In detail, on VC-Clothes, our method outperforms the previous top CC ReID method MCSC by a large improvement of 6.6\% and 5.1\% w.r.t. mAP under CC and GEN, respectively; 
In detail, on VC-Clothes, our method outperforms the previous top CC ReID method SCI by a large improvement of 2.3\% and 6.4\% w.r.t. mAP under CC and GEN, respectively;
on LaST, our method outperforms the previous top CC ReID method MCSC by a large improvement of 6.2\% under GEN; 
on Deepchange, our method outperforms the previous top CC ReID method MCSC by 1.2\% w.r.t. mAP under GEN.

\begin{table}[t]
\centering
\caption{Impact of dual prompts in our method TSPL.}
\scalebox{0.86}{
\begin{tabular}{c|c|cc|cc|cc|cc} 
\hline
\multirow{3}{*}{ADP} &\multirow{3}{*}{TAP}& \multicolumn{4}{c|}{PRCC} & \multicolumn{4}{c}{LTCC}                                                      \\ 
\cline{3-10}
            &              & \multicolumn{2}{c|}{CC}   & \multicolumn{2}{c|}{SC} & \multicolumn{2}{c|}{CC} & \multicolumn{2}{c}{SC}    \\ 
\cline{3-10}
            &              & R@1    & mAP              & R@1    & mAP            & R@1    & mAP            & R@1    & mAP              \\ 
\hline
            &              & 56.4   & 55.0             & 99.9   & 99.5           &37.2    &19.2            &84.4    &69.5           \\
\checkmark  &              & 61.1   & 57.2             & 99.5   & 99.5           & 38.8   & 18.1           & 86.3   & 73.0             \\
            % &\checkmark    &-       & -                & -      & -              & -      & -              & -      & -             \\
\checkmark  &\checkmark    & 67.0   & 60.0             & 100    & 99.5           & 52.0   & 26.3           & 88.2   & 74.2             \\
\hline
\end{tabular}
}
\label{tab:ablation}
\end{table}

\textbf{Qualitative Comparison.}\quad
\cref{fig:visual} shows some qualitative results by comparing our TSPL with the current top SC ReID method CLIP-ReID~\cite{li2023clip} and top CC ReID method FIRe\textsuperscript{2}~\cite{wang2024exploring} under SC and CC.
We can see that FIRe\textsuperscript{2} fails under SC as it does not consider the information from clothes, while CLIP-ReID mis-identifies the person with similar clothes under CC as it ignores clothes-irrelevant feature extraction.
In contrast, our TSPL works well under both SC and CC, showing effectiveness of dynamic usage of clothes information.
 
\subsection{Ablation Study}
In order to understand the individual contributions of each component in TSPL, we perform a series of ablation studies. Please note that unless explicitly mentioned, these experiments are conducted at the scale of 256$\times$128. 

\textbf{Effectiveness of Dual Prompts.}\quad 
We conduct ablation experiments by adding components one by one to analyze the contribution of appearance-decomposed prompt (ADP) and task-aware prompt (TAP) to TSPL and report the results in~\cref{tab:ablation}.
Our TSPL degrades to the baseline method CLIP-ReID, when ADP and TAP are both removed. 
When there is only ADP, our framework is designed as a dual-branch method, and the two branches correspond to the feature learning of SC and CC.
Subsequently, TAP serves as an integration strategy to incorporate the feature learning of SC and CC into a unified framework.
We can see both prompts ADP and TAP bring consistent improvements w.r.t. all metrics.
Specifically, by using ADP, we obtain a significant improvement of 4.7\% w.r.t. Rank-1 and on PRCC under CC.
It is important to note that the slight decrease in certain indicators is reasonable, as each branch was trained with data from its corresponding scenario rather than with data from all scenarios.
By further adding TAP, we gain 5.9\% and 13.2\% w.r.t. Rank-1 on PRCC and LTCC under CC respectively, which benefits from TAP integrating feature learning of different clothes scenarios into a unified framework. 
These results well demonstrate the effectiveness of both prompts, especially under CC.

\begin{table}[t]
\centering
\caption{Sensitivity of task-aware prompt. The values with green background represent the performance of the model given correct task-aware prompt at test time.}
\scalebox{0.87}{
\begin{tabular}{c|cc|cc|cc|cc} 
\hline
\multirow{3}{*}{[state]} & \multicolumn{4}{c|}{PRCC} & \multicolumn{4}{c}{LTCC}                                                      \\ 
\cline{2-9}
                      & \multicolumn{2}{c|}{CC}   & \multicolumn{2}{c|}{SC} & \multicolumn{2}{c|}{CC} & \multicolumn{2}{c}{SC}    \\ 
\cline{2-9}
                      & R@1    & mAP              & R@1    & mAP            & R@1    & mAP            & R@1    & mAP              \\ 
\hline
changed       & \cellcolor{green!20}67.0   & \cellcolor{green!20}60.0             & 100    & 99.3           & \cellcolor{green!20}52.0   & \cellcolor{green!20}26.3           & 86.1   & 69.1             \\
unchanged     & 61.4   & 55.6              & \cellcolor{green!20}100    & \cellcolor{green!20}99.5           & 48.5   & 24.5           & \cellcolor{green!20}88.2   & \cellcolor{green!20}74.2             \\
\hline
% [mask]     & -   & -              & -    & -           & -   & -           & -   & -             \\
% \hline
\end{tabular}
}
\label{tab:sensitivity}
\end{table}

\begin{figure}[t]
\begin{center}
\includegraphics[width=1.0\linewidth]{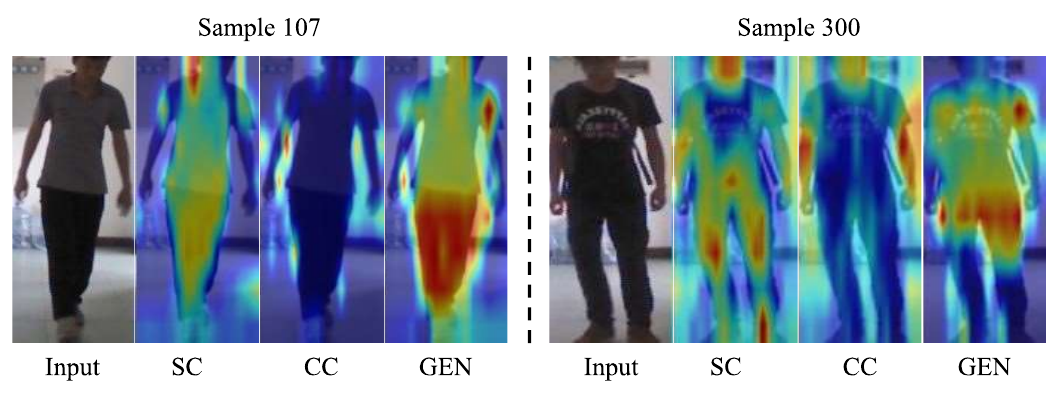}
\end{center}
  % \caption{Response maps under different clothes conditions. For the same input image, our model generates different response maps guided by the task-specific prompt.}
  \caption{Response maps under different clothes conditions. For the same input, our model produces distinct response maps guided by the task-aware prompt.}
  \label{fig:heatmap}
\end{figure}

\textbf{Task-Aware Prompt Sensitivity.}\quad 
As described in~\cref{sec:joint_training}, we employ task-aware prompts to learn scenario specific ReID features. To this end, we set [state] of the task prompt as ``changed'' for CC and as ``unchanged'' for SC, and keep consistent at training and test time. In order to verify the sensitivity of the task prompt, we implement some experiments by reversing the [state] at test time and compare with the normal experiments.
As shown in~\cref{tab:sensitivity}, the performance decreases significantly when we use the wrong [state]. For example, the performance drops by about 6\% w.r.t. Rank-1 on PRCC under CC compared with using the correct task prompt.
We also note that the performance drop of using incorrect prompts under SC is not as drastic as that under CC.
We attribute this phenomenon to two primary factors: firstly, the simplicity of the SC within the dataset; 
secondly, it is reasonable to use the clothes irrelevant features of CC prompts under SC, but it ignores the importance of clothes information in SC and makes some cases that originally rely on clothes information retrieval difficult.
In fact, using incorrect task prompts can be avoided in practice as we ask the user to give a proper prompt according to the current scenario; when the user is uncertain about the clothes condition, one can switch to the GEN mode, for which we also provide better results than previous methods.
% We also provide visualization of response maps in~\cref{fig:heatmap} to observe how the task-specific prompt adjusts image features.
% These results indicate that task-specific prompts indeed guide customized features learning for different scenarios.
We also visualize response maps in \cref{fig:heatmap} to show how task-aware prompts modulate image features. These results confirm that such prompts effectively guide feature learning across different scenarios.

\textbf{Appearance-Decomposed Prompt Design for CC.}\quad
For CC ReID, one intuitive way of using the appearance-decomposed prompt is to directly mask out the clothes-relevant information through the 0-1 mask mechanism in Natural Language Processing~\cite{Yoav} and only rely on identity-relevant features for ReID. Yet, we claim that clothes may also contain useful information for CC ReID but in a summarized way. Thus, we retain the potentially beneficial information in clothes by averaging the prompts embeddings from the same person with different clothes. We compare the results of the above two strategies in~\cref{tab:adp-design}, where we can see our averaging strategy outperforms the masking out one by 2.1\% and 4\% w.r.t. Rank-1 on PRCC and LTCC, respectively.
% These results suggest that moderately retaining common clothes knowledge about persons by averaging is beneficial for retrieval.
These results suggest that moderately averaging common clothes knowledge benefits retrieval.

\begin{table}
\centering
\caption{Appearance-decomposed prompt design for the CC scenario. None means no operation is performed.}
\begin{tabular}{c|cc|cc} 
\hline
\multirow{2}{*}{ADP Design} & \multicolumn{2}{c|}{PRCC} & \multicolumn{2}{c}{LTCC}                                                      \\ 
\cline{2-5}
                           & R@1    & mAP           & R@1    & mAP              \\ 
\hline
None                       & 57.6   & 46.8          & 40.1   & 18.6            \\
\hline
Masking                    & 64.9   & 57.9          & 48.0   & 24.6            \\
Averaging                  & 67.0   & 60.0          & 52.0   & 26.3            \\
\hline
\end{tabular}
\label{tab:adp-design}
\end{table}

\textbf{Visualization of Response Maps.} \quad
We visualize the response maps using the GradCAM method\footnote{\url{https://github.com/jacobgil/pytorch-grad-cam}} to observe how the task-aware prompt adjusts image features.
As shown in~\cref{fig:heatmap}, for the same image, when using the task-aware prompt of SC, we observe high responses on clothes area. In contrast, when employing the task-aware prompt of CC, we observe low responses on clothes area while high responses on human body contour. The response under the task-aware prompt of GEN can be seen as the result of a comprehensive consideration of SC and CC.
For example, sample 107 in modest clothes-changed dataset PRCC, which only the top changes while the pants remain the same. While focusing on other information, the feature under the GEN prompt responds most strongly to the pants; moreover, the features prompted by SC and CC also have different focuses, SC focuses more on clothes parts, while CC emphasizes non-clothes parts. 
These results confirm that task-aware prompt can effectively modulate image features.

\begin{figure}[t]
\begin{center}  
\includegraphics[width=1.0\linewidth]{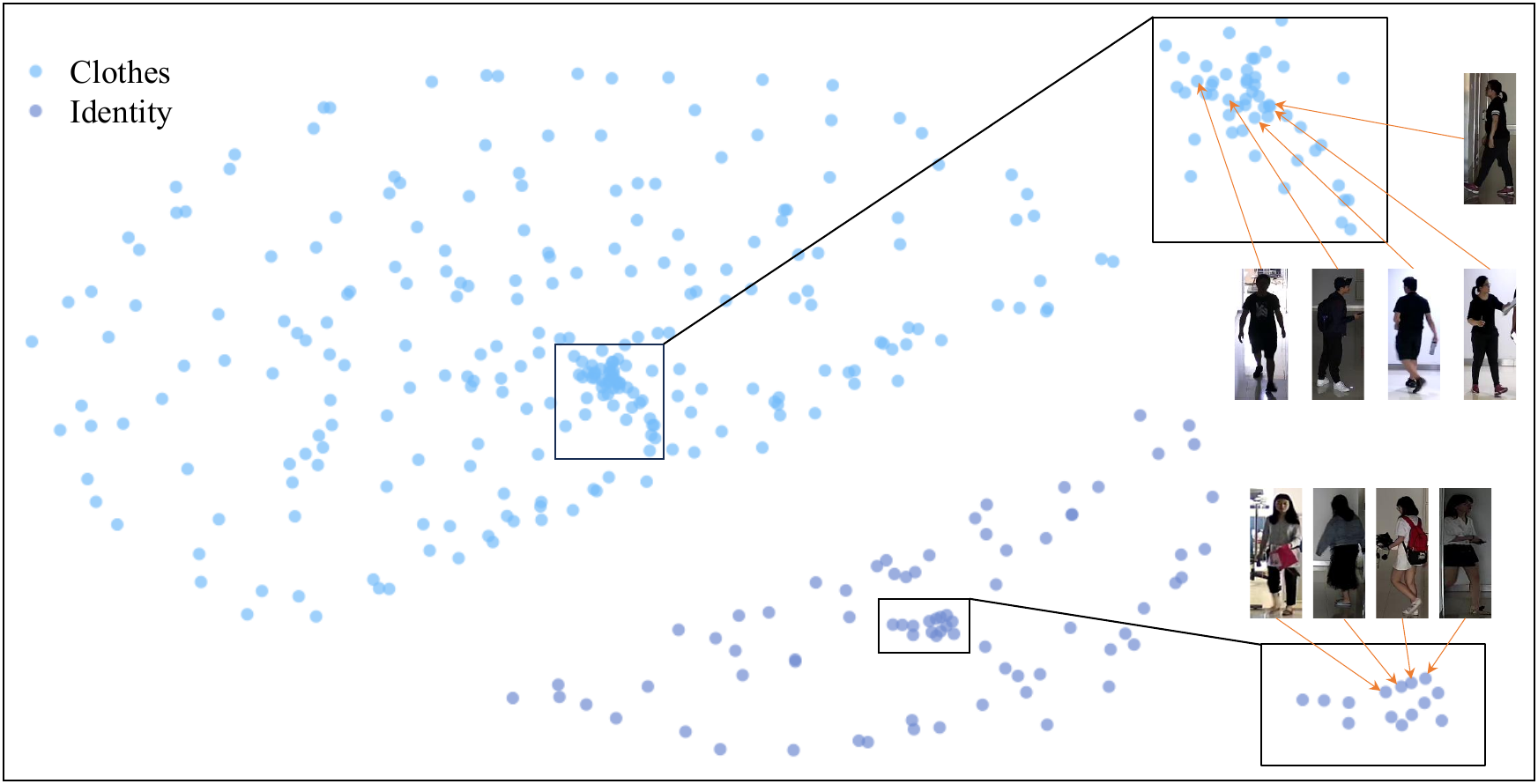}
\end{center}
  \caption{Embedding distribution of clothes and identity learnable context vector. Please zoom in for better view.}
  \label{fig:vector}
\end{figure}

\textbf{Analysis of Learnable Clothes and Identity  Context Vector.}\quad 
% For their corresponding learnable context vectors in the embedding space, 
% previous research~\cite{zhou2022learning,zhou2022conditional} has shown that ``\textit{the semantics of the vectors is not necessarily correlated with the nearest words}''.
% Therefore, it is impractical to directly interpret the learnable clothes and identity context vector. 
The pseudo-word has no actual meaning. For their corresponding learnable context vectors in the embedding space, it is infeasible to associate them with existing pretrained word embeddings to interpret their semantic meaning, since they are learned independently~\cite{zhou2022learning,zhou2022conditional}.
Therefore, we resort to visualize their distributions through t-SNE~\cite{van2008visualizing} to prove that learnable clothes and identity context vectors focus on different information.
As shown in~\cref{fig:vector}, the distributions of the learnable clothes and identity context vectors are clearly separated, indicating a clear distinction between them in the embedding space. For clothes embedding distribution, most learnable vectors maintain a suitable distance, while some vectors are clustered due to similar clothes, such as black clothes. For identity embedding distribution, a similar phenomenon is observed, with some vectors clustered based on clothes-irrelevant information, such as gender, long hair, and posture.
This demonstrates that the learnable clothes and identity context vectors indeed capture clothes- and identity-relevant information, respectively, thereby decoupling clothes and identity information.

\textbf{Task-Aware Prompt Design.} \quad
To demonstrate the effectiveness of our task prompt design, we compare it with the instruction-based approach of Instruct-ReID~\cite{he2023instructreid}, which unifies different ReID tasks through task-aware instructions. Following their settings, we adopt the instructions for SC (\textit{Do not change clothes}) and CC (\textit{With clothes changed}) to train our TSPL and evaluate it on PRCC and LTCC. As shown in \cref{tab:compare-tsp}, our method not only generalizes to arbitrary clothes scenarios but also achieves superior performance. Note that Instruct-ReID does not report GEN results, as it dose not have prompt for GEN task.

\begin{table}
\centering
\caption{Different task prompts design. Our method covers arbitrary clothes scenarios and achieves better performance.
} 
\begin{tabular}{l|cc|cc|cc} 
\hline
\multirow{3}{*}{Task Prompt}           & \multicolumn{6}{c}{PRCC}  \\
\cline{2-7}
                                   &\multicolumn{2}{c|}{CC} & \multicolumn{2}{c|}{SC} & \multicolumn{2}{c}{GEN} \\
\cline{2-7}
                                   &R@1   &mAP   &R@1   &mAP   &R@1   &mAP \\ 
\hline
Instruct-ReID's                    &59.9  &59.1  &100   &99.3  &-     &- \\
\hline
Ours                               &67.0  &60.0  &100   &99.5  &99.8  &82.3  \\
\hline
\hline
\multirow{3}{*}{Task Prompt}           & \multicolumn{6}{c}{LTCC}  \\
\cline{2-7}
                                   &\multicolumn{2}{c|}{CC} & \multicolumn{2}{c|}{SC} & \multicolumn{2}{c}{GEN} \\
\cline{2-7}
                                   &R@1   &mAP   &R@1   &mAP   &R@1   &mAP \\ 
\hline
Instruct-ReID's                    &44.9  &22.7  &85.9  &69.4  &-     &- \\
\hline
Ours                               &52.0  &26.3  &88.2  &74.2  &81.1  &48.0 \\
\hline
\end{tabular}
\label{tab:compare-tsp}
\end{table}

\textbf{Different Sampling Rules for CC.} \quad
Typical \textit{P-K} sampling balances the long-tailed distribution of ID samples by randomly selecting a fixed number of \textit{K} images for each ID within a minibatch. However, it only considers ID variations and ignores clothes variations.
To better associate the SC and CC task prompts with clothes changes while mitigating the long-tailed distribution of clothes per ID, we incorporate clothes variations into the sampling rule under CC during training. 
As mentioned in Section IV-B, based on \textit{P-K} sampling, we specify that each ID in a minibatch under CC has \textit{M} clothes, with \textit{K}/\textit{M} samples per clothes. Due to the inconsistent number of clothes per ID, we set \textit{M}=2 in our experiments to balance the clothes variations of each ID in a minibatch under CC. 
Note that the \textit{M}=2 clothes are randomly chosen for each minibatch, rather than fixed across all batches. As training progresses, the sampling process gradually covers all clothes associated with each ID.
We report the performance comparison of our method using different sampling rules under CC in~\cref{tab:sampling}. Here, when \textit{M}=1, it means that the samples of ID in each minibatch under CC belong to different clothes. It can be seen that our sampling rule with reasonable clothes variations significantly outperforms the typical \textit{P-K} sampling under CC.

\begin{table}[t]
\centering
\caption{With different sampling rules for CC, introducing clothes variations allows the CC task prompt to better modulate features toward the CC, leading to improved performance.}
\begin{tabular}{c|cc|cc} 
\hline
\multirow{2}{*}{Sampling Rule} & \multicolumn{2}{c|}{PRCC} & \multicolumn{2}{c}{LTCC}                                                      \\ 
\cline{2-5}
                            & R@1    & mAP            & R@1    & mAP              \\ 
\hline
P-K\cite{Luo2020}           & 56.7   & 56.1           & 39.5   & 22.4         \\
\hline
\textit{M}=1                & 63.4   & 57.0           & 45.9   & 23.0         \\
\textit{M}=2 (Ours)         & 67.0   & 60.0           & 52.0   & 26.3            \\
\hline   
\end{tabular}
\label{tab:sampling}
\end{table}

\subsection{Analysis of Hyper-parameter}
\textbf{Number of Learnable Context Vectors $u$ and $v$.}\quad
We analyze the parameters $u$ and $v$ of the learnable context vectors $\{c_1, c_2, \cdots, c_u\}$ and $\{h_1, h_2, \cdots, h_v\}$ assigned to the pseudo-words {clothes} and {identity} in the ADP. To narrow the parameter search space, we set $u$ and $v$ to be equal.
Following CLIP-ReID~\cite{li2023clip}, we choose $1$, $4$ and $8$ for $u$ and $v$, and report the results in~\cref{tab:uv}.
We find that setting $u$ and $v$ to $1$ fails to learn sufficient text descriptions, but when the length increases to $8$, it is seem redundant and unhelpful. 
Best results are achieved on both PRCC and LTCC when set $u$ and $v$ to 4.
Therefore, we finally set $u$ and $v$ to $4$ in all experiments.

\begin{table}
\centering
\caption{Analysis of the number of learnable context vectors $u$ and $v$ for appearance-decomposed prompt.}
\scalebox{0.92}{
\begin{tabular}{c|cc|cc|cc|cc} 
\hline
\multirow{3}{*}{$u$, $v$} & \multicolumn{4}{c|}{PRCC} & \multicolumn{4}{c}{LTCC}                                                      \\ 
\cline{2-9}
                           & \multicolumn{2}{c|}{CC}   & \multicolumn{2}{c|}{SC} & \multicolumn{2}{c|}{CC} & \multicolumn{2}{c}{SC}    \\ 
\cline{2-9}
                           & R@1    & mAP              & R@1    & mAP            & R@1    & mAP            & R@1    & mAP              \\ 
\hline
1                          & 65.8   & 50.8             & 99.9   & 98.2           & 46.4   & 23.2           & 87.8   & 68.3             \\
4(Ours)                    & 67.0   & 60.0             & 100    & 99.5           & 52.0   & 26.3           & 88.2   & 74.2             \\
8                          & 64.3   & 50.8             & 99.9   & 97.7           & 49.0   & 23.1           & 86.8   & 68.1             \\

\hline
\end{tabular}
}
\label{tab:uv}
\end{table}

\begin{table}[t]
\centering
\caption{Analysis of the number of learnable context vectors $w$ for task-aware prompt.}
\scalebox{0.92}{
\begin{tabular}{c|cc|cc|cc|cc} 
\hline
\multirow{3}{*}{$w$} & \multicolumn{4}{c|}{PRCC} & \multicolumn{4}{c}{LTCC}                                                      \\ 
\cline{2-9}
                           & \multicolumn{2}{c|}{CC}   & \multicolumn{2}{c|}{SC} & \multicolumn{2}{c|}{CC} & \multicolumn{2}{c}{SC}    \\ 
\cline{2-9}
                           & R@1    & mAP              & R@1    & mAP            & R@1    & mAP            & R@1    & mAP              \\ 
\hline
0                          & 65.5   & 58.7             & 100    & 99.4           & 48.2   & 24.1           & 88.2   & 74.1             \\
1                          & 64.9   & 48.8             & 99.9   & 97.7           & 46.4   & 22.1           & 85.6   & 66.2             \\
4(Ours)                    & 67.0   & 60.0             & 100    & 99.5           & 52.0   & 26.3           & 88.2   & 74.2             \\
8                          & 63.7   & 45.9             & 99.8   & 96.7           & 46.9   & 22.7           & 87.1   & 67.3             \\

\hline
\end{tabular}
}
\label{tab:w}
\end{table}

\begin{table}[t]
\centering
\caption{Comparison on Market1501 and MSMT17.}
\begin{tabular}{c|cc|cc} 
\hline
\multirow{2}{*}{-} & \multicolumn{2}{c|}{Market1501} & \multicolumn{2}{c}{MSMT17}  \\ 
\cline{2-5}
                                          & Rank-1    & mAP        & Rank-1    & mAP              \\ 
\hline
FSAM~\cite{hong2021fine}                  & 94.6      & 85.6       & -         & -                \\
CAL~\cite{Gu2022}                         & 94.7      & 87.5       & 79.7      & 57.3             \\
Instruct-ReID~\cite{he2023instructreid}   & 96.2      & 92.3       & 86.2      & 71.9             \\
\hline
CLIP-ReID~\cite{li2023clip}               & 95.4      & 90.5       & 89.7      & 75.8             \\
Ours                                      & 95.3      & 89.3       & 88.7      & 74.6             \\
\hline
\end{tabular}
\label{tab:market-msmt}
\end{table}

\textbf{Number of Learnable Context Vectors $w$.} \quad
We study the configuration of learnable context vectors ${s_1, s_2, \cdots, s_w}$ assigned to a pseudo-word ${s}$ in a task-aware prompt within the embedding space, and analyze the impact of varying the number of vectors $w$. 
\cref{tab:w} summarizes the results of the analysis. 
We observe that it is best to set $w$ to $4$ for PRCC and LTCC.
Therefore, we set $w=4$ in all experiments.

\subsection{Generalization Discussion}
\textbf{Results on Clothes Consistent Datasets.} \quad
To demonstrate the generalizability of our method, we report its performance on traditional clothes-consistent datasets such as Market-1501~\cite{zheng2015scalable} and MSMT17~\cite{wei2018person}. Since there is no clothes variation in clothes consistent datasets, our method degenerates to a method similar to CLIP-ReID. Thus, as shown in~\cref{tab:market-msmt}, the results on clothes consistent datasets are approximately the same as those reported in CLIP-ReID.
Meanwhile, our method outperforms other CC ReID methods such as CAL by a large margin on the challenging dataset MSMT17.

\section{Conclusion}
To address the ReID task under different clothes conditions in real-world scenarios, in this paper, we propose a unified framework, Two-Stage Prompt Learning (TSPL), which integrates \textit{clothes–identity decoupling} and \textit{scenario-aware task prompts} within VLMs.
In Stage 1, appearance-decomposed prompt learning disentangles clothes and identity information under frozen encoders, producing stable embeddings where clothes cues can be selectively preserved or suppressed. In Stage 2, task-aware prompt learning trains the image encoder with learnable context embeddings guided by Stage-1 representations, enabling robust adaptation across SC, CC, and GEN conditions. 
Extensive experiments on PRCC, LTCC, VC-Clothes, LaST, and DeepChange demonstrate the effectiveness of our TSPL in addressing ReID under different clothes conditions within a unified framework.

\section{Limitation and future work}
\begin{figure}[t]
\begin{center}  
\includegraphics[width=1.0\linewidth]{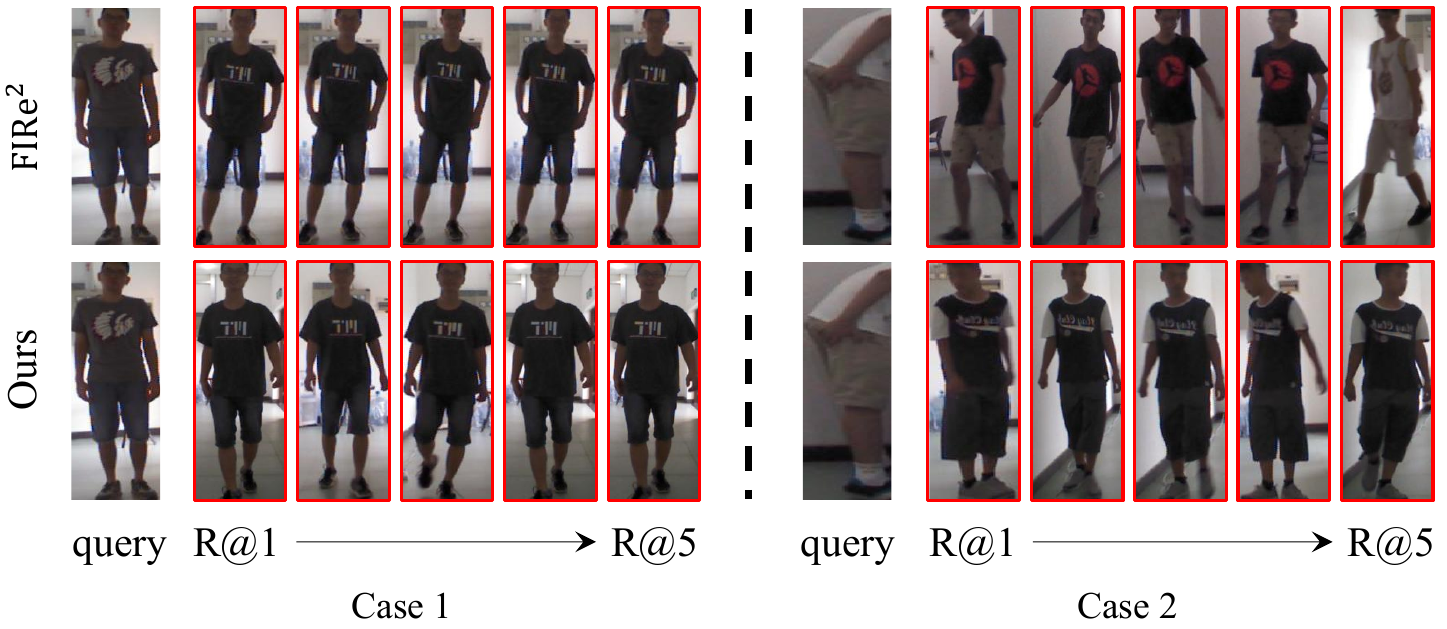}
\end{center}
  \caption{Failure cases under CC: (Case 1) similar body shapes; (Case 2) severe occlusion.}
  \label{fig:failure_case}
\end{figure}

Despite our method achieved state-of-the-art results, several limitations remain. As shown in~\cref{fig:failure_case},  when different identities share highly similar body structures, both our method and existing approaches struggle to distinguish them effectively. Although our framework mitigates clothes variation, it still relies on relatively coarse identity cues. Learning more fine-grained discriminative information remains an open problem.
Moreover, under heavy occlusion, critical identity regions are missing, leading to performance degradation for both methods. Since our framework does not explicitly model occlusion or part-level reconstruction, it is inherently limited in such cases.

In future work, more fine-grained identity modeling strategies could serve as potential directions for improvement, such as optimizing feature extraction under different scenarios during training or leveraging auxiliary cues (e.g., silhouette or human parsing information) to enhance robustness and generalization ability.

\bibliographystyle{IEEEtran}
\bibliography{main}

\vfill

\end{document}